\documentclass[runningheads]{llncs}

\usepackage[T1]{fontenc}
\usepackage{lmodern}
\usepackage{textcomp}
\usepackage{float}

\usepackage{graphicx}
\usepackage{url}

\usepackage{bm}

\usepackage{booktabs}
\begin{document}

\title{Distinguishing between roles of football\\ players in play-by-play match event data}
\titlerunning{Distinguishing between roles of football players}

\author{Bart Aalbers \and Jan Van Haaren}
\authorrunning{B. J. Aalbers and J. Van Haaren}

\institute{SciSports, Hengelosestraat 500, 7521 AN Enschede, Netherlands\\\email{\{b.aalbers, j.vanhaaren\}@scisports.com}}

\maketitle

\begin{abstract}
Over the last few decades, the player recruitment process in professional football has evolved into a multi-billion industry and has thus become of vital importance. To gain insights into the general level of their candidate reinforcements, many professional football clubs have access to extensive video footage and advanced statistics. However, the question whether a given player would fit the team's playing style often still remains unanswered. In this paper, we aim to bridge that gap by proposing a set of 21 player roles and introducing a method for automatically identifying the most applicable roles for each player from play-by-play event data collected during matches.

\end{abstract}

\section{Introduction}
\label{sec:introduction}

Over the last few decades, professional football has turned into a multi-billion industry. During the 2017/2018 season, the total revenue of all twenty English Premier League clubs combined exceeded a record high of 5.3 billion euro with 1.8 billion euro spent on the acquisition of players in the 2017 summer transfer window alone~\cite{deloitte2018annual,cies2018seas}. Hence, the player recruitment process has become of vital importance to professional football clubs. That is, football clubs who fail to strengthen their squads with the right players put their futures at stake.

Compared to twenty years ago, scouts at football clubs have plenty more tools at their disposal to find appropriate players for their clubs. Scouts nowadays can watch extensive video footage for a large number of players via platforms like Wyscout\footnote{\url{https://www.wyscout.com}} and obtain access to advanced statistics, performance metrics and player rankings via platforms like SciSports Insight.\footnote{\url{https://insight.scisports.com}} Although these tools provide many insights into the abilities and general level of a player, they largely fail to answer the question whether the player would fit a certain team's playing style. For example, star players like Cristiano Ronaldo and Lionel Messi operate at a comparable level but clearly act and behave in different ways on the pitch.

To help answer this question, we compose a set of 21 candidate player roles in consultation with the SciSports Datascouting department and introduce an approach to automatically identify the most applicable roles for each player from play-by-play event data collected during matches. To optimally leverage the available domain knowledge, we adopt a supervised learning approach and pose this task as a probabilistic classification task. For each player, we first extract a set of candidate features from the match event data and then perform a separate classification task for each of the candidate player roles.

\section{Dataset}
\label{section:data}

Our dataset consists of play-by-play match event data for the 2017/2018 season of the English Premier League, Spanish LaLiga, German 1. Bundesliga, Italian Serie A, French Ligue Un, Dutch Eredivisie and Belgian Pro League provided by SciSports' partner Wyscout. Our dataset represents each match as a sequence of roughly 1500 on-the-ball events such as shots, passes, crosses, tackles, and interceptions. For each event, our dataset contains a reference to the team, a reference to the player, the type of the event (e.g., a pass), the subtype of the event (e.g., a cross), 59 boolean indicators providing additional context (e.g., accurate or inaccurate and left foot or right foot), the start location on the pitch and when relevant also the end location on the pitch.

\section{Player roles}
\label{section:player-roles}

\begin{figure}[htp]
  \begin{center}
    \includegraphics[width=\textwidth]{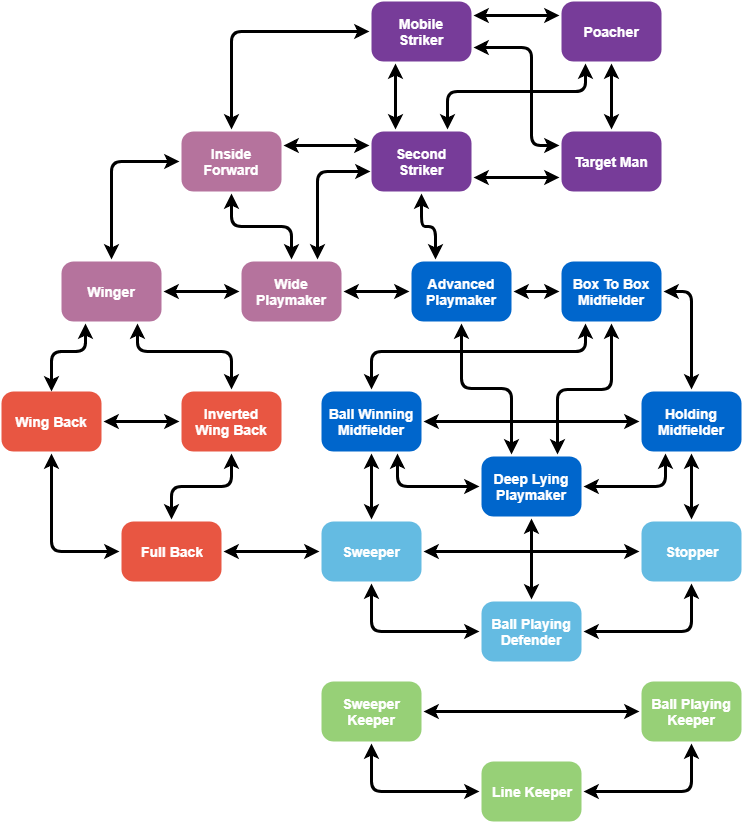}
    \caption{A graphical overview of the 21 proposed player roles for six different positions. Roles that have tasks and duties in common are connected by arrows. Roles that apply to the same position are in the same color. Roles for goalkeepers are in green, roles for defenders are in light blue, roles for backs are in red, roles for central midfielders are in dark blue, roles for wingers are in pink and roles for forwards are in purple.}
    \label{fig:player_roles}
  \end{center}
\end{figure}

We compile a list of 21 player roles based on an extensive sports-media search (e.g.,~\cite{arse2016dm,espn2018dm,thompson2018four}) and the Football Manager 2018 football game, which is often renowned for its realism and used by football clubs in their player recruitment process (e.g.,~\cite{goal2012how,sullivan2016beautiful}). In consultation with the football experts from the SciSports Datascouting department, we refine the obtained list and also introduce two additional roles, which are the mobile striker and the ball-playing goalkeeper. The mobile striker is dynamic, quick and aims to exploit the space behind the defensive line. The ball-playing goalkeeper has excellent ball control and passing abilities and often functions as an outlet for defenders under pressure.

For each of these roles, we define a set of key characteristics that can be derived from play-by-play match event data. Most of the defined player roles describe players who excel in technical ability, game intelligence, strength, pace or endurance. Figure~\ref{fig:player_roles} presents an overview of the 21 player roles, where related roles are connected by arrows and roles for the same position are in the same color. For example, a full back and wing back have similar defensive duties, which are marking the opposite wingers, recovering the ball, and preventing crosses and through balls. In contrast, a wing back and winger have similar offensive duties, which are providing width to the team, attempting to dribble past the opposite backs, and creating chances by crossing the ball into the opposite box.

Due to space constraints, we focus this paper on the five roles for central midfielders. We now present each of these five roles in further detail.

\begin{description}
  \item[Ball-Winning Midfielder (BWM):] mainly tasked with gaining possession. When the opponent has possession, this player is actively and aggressively defending by closing down opponents and cutting off pass supply lines. This player is a master in disturbing the build-up of the opposing team, occasionally making strategic fouls such that his team can reorganize. When in possession or after gaining possession, this player plays a simple passing game. A player in this role heavily relies on his endurance and game intelligence. Notable examples for this role are Kant\'{e} (Chelsea), Casemiro (Real Madrid) and Lo Celso (Paris Saint-Germain).
  \item[Holding Midfielder (HM):] tasked with protecting the defensive line. When the opponent has possession, this player defends passively, keeps the defensive line compact, reduces space in front of the defence and shadows the opposite attacking midfielders. In possession, this player dictates the pace of the game. This role requires mostly game intelligence and strength. Notable examples for this role are Busquets (Barcelona), Weigl (Borussia Dortmund) and Mati\'{c} (Manchester United).
  \item[Deep-Lying Playmaker (DLP):] tasked with dictating the pace of the game, creating chances and exploiting the space in front of his team's defense. This player has excellent vision and timing, is technically gifted and has accurate passing skills to potentially cover longer distances as well. Hence, this player is rather focused on build-up than defense. He heavily relies on his technical ability and game intelligence. Notable examples for this role are Jorginho (transferred from Napoli to Chelsea), F\`{a}bregas (Chelsea) and Xhaka (Arsenal).
  \item[Box-To-Box (BTB):] a more dynamic midfielder, whose main focus is on excellent positioning, both defensively and offensively. When not in possession, he concentrates on breaking up play and guarding the defensive line. When in possession, this player dribbles forward passing the ball to players higher up the pitch and often arrives late in the opposite box to create a chance. This player heavily relies on endurance. Notable examples for this role are Wijnaldum (Liverpool), Matuidi (Juventus) and Vidal (transferred from Bayern M\"{u}nchen to Barcelona).
  \item[Advanced Playmaker (AP):] the prime creator of the team, occupying space between the opposite midfield and defensive line. This player is technically skilled, has a good passing range, can hold up a ball and has excellent vision and timing. He relies on his technical ability and game intelligence to put other players in good scoring positions by pinpointing passes and perfectly timed through balls. Notable examples for this role are De Bruyne (Manchester City), Luis Alberto (Lazio Roma) and Coutinho (Barcelona).
\end{description}

\section{Approach}
\label{section:approach}

In this section, we propose a method to automatically derive the most applicable player roles for each player from play-by-play match event data. We adopt a supervised learning approach to optimally leverage the available domain knowledge within the SciSports Datascouting department. More specifically, we address this task by performing a probabilistic classification task for each of the candidate roles introduced in Section~\ref{section:player-roles}. In the remainder of this section, we explain the feature engineering and probabilistic classification stages in more detail.

\subsection{Feature engineering}
\label{sub:ftrc}

The goal of the feature engineering stage is to obtain a set of features that both characterizes the candidate player roles and distinguishes between them. The most important challenges are to distinguish between the qualities and quantities of a player's actions and to account for the strength of a player's teammates and opponents as well as the tactics and strategies employed by a player's team.

We perform the feature engineering stage in two steps. In the first step, we compute a set of basic statistics and more advanced performance metrics. We base these statistics and metrics on the specific tasks players in the different roles perform. We normalize these basic statistics and metrics in two different ways to improve their comparability across players. In the second step, we standardize the obtained statistics and metrics to facilitate the learning process and construct features by combining them based on domain knowledge.

\subsubsection{Computing statistics and metrics.}

We compute a set of 242 basic statistics and metrics for each player across a set of matches (e.g., one full season of matches). This set includes statistics like the number of saves from close range, the number of duels on the own half, the number of passes given into the opposite box, the number of long balls received in the final third, the number of offensive ground duels on the flank and the number of high-quality attempts. For example, a ball-winning midfielder is tasked with regaining possession, aggressively pressing opponents and laying off simple passes to more creative teammates. Hence, we compute statistics like the number of interceptions in the possession zone (i.e., the middle of the pitch), the number of defensive duels won in the possession zone, and the number of low-risk passes from the possession zone.

We normalize these statistics and metrics in two different ways to obtain 484 features. First, we normalize the values by the total number of actions performed by the player. This normalization helps to improve the comparability across players on strong and weak teams, where strong teams tend to perform more on-the-ball actions. Second, we normalize the values by the total number of actions performed by the player's team. This normalization helps to identify key players that are often in possession of the ball (e.g., playmakers).

In addition, we compute 31 metrics including centrality measures in the passing network, defensive presence in the team and a player's wide contribution.

\subsubsection{Constructing features.}

We construct a set of 181 key features by combining the statistics and metrics. In order to facilitate the learning process, we standardize our set of 515 features. More specifically, we project the feature values to the $[\mu-2\sigma, \mu+2\sigma]$ interval, where $\mu = 0$ and $\sigma = 1$. We project any extreme values that fall outside the interval to the interval bounds.

We use the resulting set of standardized features to construct the key feature set. For example, we combine the statistics representing the number of attempts from the area immediately in front of goal and the expected-goals metrics for these attempts into a key feature reflecting the frequency of high-quality attempts from short range.

\subsection{Probabilistic classification}
\label{sub:modc}

We obtain labels for a portion of our dataset from the SciSports Datascouting department. In particular, we request the Datascouting department to assign the primary role to each of the players in our dataset they feel confident enough about to assess. Although football players often fulfill several different roles during the course of a season or even within a single match, we only collect each player's primary role to simplify our learning setting.

We perform a separate binary classification task for each of the 21 roles. For each role, we construct a separate dataset in two steps. First, we obtain all examples of players that were assigned that role and label them as positive examples. Second, we obtain all examples of players that were assigned one of the ``disconnected'' roles in Figure~\ref{fig:player_roles} as well as 25\% of the examples for players that were assigned one of the ``connected'' roles and label them as negative examples.

We notice that keeping 25\% of the examples for players having a connected role improves the accuracy of our classification models. If we keep all connected examples, the learner focuses on the specific tasks that distinguish between the role of interest and the connected roles. If we remove all connected examples, the learner focuses on distinguishing between positions rather than specific roles.

\begin{figure}[!htp]
  \vspace{-10pt}
  \begin{center}
    \includegraphics[width=\textwidth]{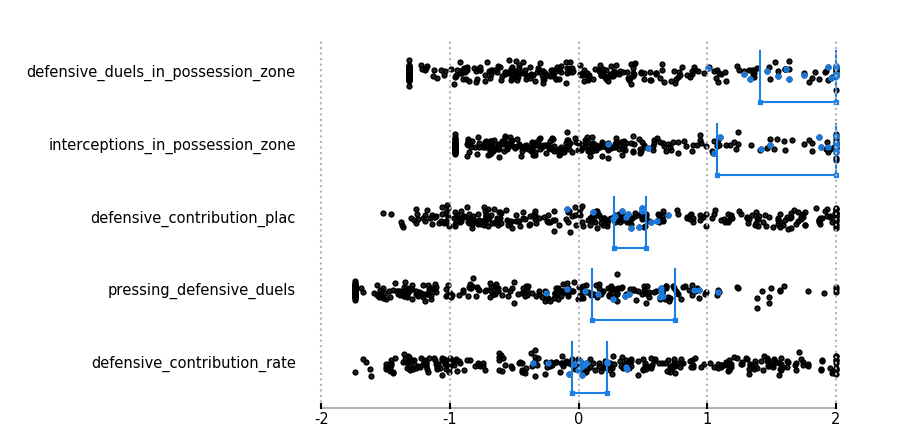}
    \caption{The five most relevant features for ball-winning midfielders. The blue dots represent players labeled as ball-winning midfielders, while the black dots represent players labeled as any of the other roles. The blue lines represent the optimal region.}
    \label{fig:norm_feat}
  \end{center}
  \vspace{-20pt}
\end{figure}

\begin{figure}[!htp]
  \vspace{-10pt}
  \begin{center}
    \includegraphics[width=\textwidth]{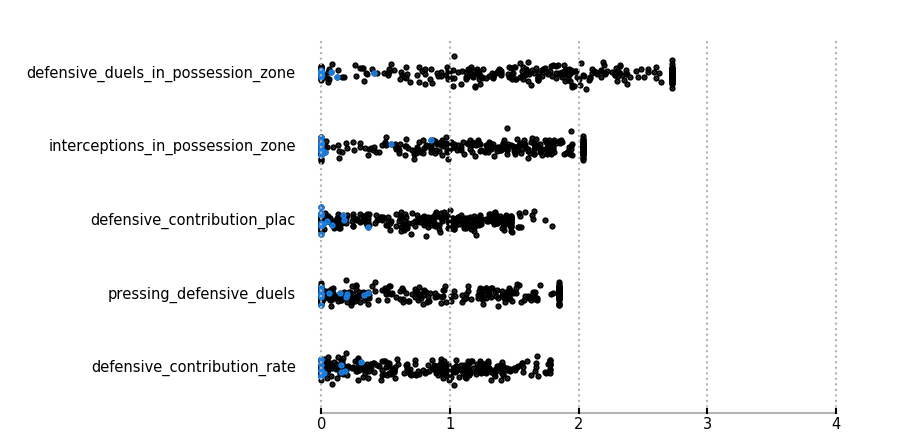}
    \caption{The five most relevant features for ball-winning midfielders after the transformation. The blue dots represent players labeled as ball-winning midfielders, while the black dots represent players labeled as any of the other roles. The optimal value is 0.}
    \label{fig:dist_feat}
  \end{center}
  \vspace{-20pt}
\end{figure}

We perform a final transformation of the key feature values before they are fed into the classification algorithm. For each key feature and each role, we first define the optimal feature value range and then compute the distance between the actual feature value and the optimal range. We determine the optimal range by computing percentile bounds on the labeled training data. We set a parameter $\beta$, which translates into a lower bound of $\beta$ and an upper bound of $1-\beta$, as can be seen in Figure~\ref{fig:norm_feat}. For example, $\beta = 0.25$ translates into the $[0.25,0.75]$ range. We set the feature values that fall within the optimal range to 0 and all other feature values to their distance to the closest bound of the optimal range. Hence, we project the original $[-2,2]$ range to a $[0,4]$ range, where $0$ is the optimal value, as can be seen in Figure~\ref{fig:dist_feat}.

\section{Experimental evaluation}

This section presents an experimental evaluation of our approach. We present the methodology, formulate the research questions and discuss the results.

\subsection{Methodology}

We construct one example for each player in our dataset. After dropping all players who played fewer than 900 minutes, we obtain a set of 1910 examples. The SciSports Datascouting department manually labeled 356 of these examples (i.e., 18.6\% of the examples) as one of the 21 roles introduced in Section~\ref{section:player-roles}.

We prefer algorithms learning interpretable models to facilitate the explanation of the inner workings of our approach to football coaches and scouts. More specifically, we train a Stochastic Gradient Descent classifier using logistic loss as the objective function for each role. Hence, the coefficients of the model reflect the importances of the features.

For each role, we perform ten-fold cross-validation and use the Synthetic Minority Over-Sampling Technique (SMOTE) to correct the balance between the positive and negative examples~\cite{chaw2002smote}. After performing the transformation explained in Section~\ref{sub:modc}, we obtain a set of over 300 examples.

\subsection{Discussion of results}

In this section, we investigate what the optimal values for the regularization parameter $\alpha$ and boundary parameter $\beta$ are and who the most-suited players are to fulfill each of the central midfield roles.

\subsubsection{What are the optimal values for $\alpha$ and $\beta$?}
\label{sub:optim}

We need to optimize the regularization parameter $\alpha$ and the boundary parameter $\beta$ to obtain the most-relevant features for each of the roles. We try a reasonable range of values for both parameters and optimize the weighted logistic loss due to the class imbalance.

\begin{table}
\centering
\caption{The weighted logistic loss across the roles and the folds for different values of the regularization parameter $\alpha$ and the boundary parameter $\beta$. We obtain the optimal weighted logistic loss for $\alpha = 0.050$ and $\beta = 0.250$. The best result is in bold.}
\label{table:optimization}
\begin{tabular}{rrr}
\toprule
\bm{$\alpha$} & \bm{$\beta$} & \textbf{Weighted logistic loss} \tabularnewline
\midrule
1.000 & 0.250 & 0.2469 \tabularnewline
0.500 & 0.250 & 0.1800 \tabularnewline
0.100 & 0.250 & 0.0977 \tabularnewline
\textbf{0.050} & \textbf{0.250} & \textbf{0.0831} \tabularnewline
0.010 & 0.250 & 0.0835 \tabularnewline
0.005 & 0.250 & 0.1034 \tabularnewline
0.001 & 0.250 & 0.2199 \tabularnewline
\midrule
0.050 & 0.100 & 0.0865 \tabularnewline
0.050 & 0.200 & 0.0833 \tabularnewline
\textbf{0.050} & \textbf{0.250} & \textbf{0.0831} \tabularnewline
0.050 & 0.300 & 0.0840 \tabularnewline
0.050 & 0.350 & 0.0863 \tabularnewline
0.050 & 0.400 & 0.0890 \tabularnewline
\bottomrule
\end{tabular}
\end{table}

Table~\ref{table:optimization} shows the weighted logistic loss across the ten folds and 21 roles for several different values for the regularization parameter $\alpha$ and the boundary parameter $\beta$. The top half of the table shows that 0.050 is the optimal value for the regularization parameter $\alpha$, while the bottom half of the table shows that 0.250 is the optimal value for the boundary parameter $\beta$.

\subsubsection{What are the most suitable players for the central midfielder roles?}

\begin{table}
\centering
\caption{The probabilities for the five central midfielder roles for 20 players in our dataset. For each role, we show the top-ranked labeled player as well as the top-three-ranked unlabeled players. The highest probability for each player is in bold.}
\label{table:results}
\begin{tabular}{lccccccc}
\toprule
\textbf{Player} & \textbf{Labeled} & \textbf{BWM} & \textbf{HM} & \textbf{DLP} & \textbf{BTB} & \textbf{AP}\tabularnewline
\midrule
Trigueros & Yes & \textbf{0.98} & 0.85 & 0.87 & 0.89 & 0.29\tabularnewline
Sergi Darder & No & \textbf{0.98} & 0.90 & 0.82 & 0.94 & 0.11\tabularnewline
Guilherme & No & \textbf{0.97} & 0.94 & 0.32 & 0.93 & 0.03\tabularnewline
E. Skhiri & No &\textbf{0.97} & 0.93 & 0.38 & 0.96 & 0.04\tabularnewline
\midrule
N. Matic & Yes & 0.91 & \textbf{0.98} & 0.82 & 0.42 & 0.01\tabularnewline
J. Guilavogui & No & 0.96 & \textbf{0.97} & 0.71 & 0.76 & 0.05\tabularnewline
Rodrigo & No & 0.90 & \textbf{0.96} & 0.23 & 0.48 & 0.00\tabularnewline
G. Pizarro & No & 0.90 & \textbf{0.96} & 0.24 & 0.41 & 0.00\tabularnewline
\midrule
Jo\~{a}o Moutinho & Yes & 0.91 & 0.88 & \textbf{0.98} & 0.41 & 0.14\tabularnewline
J. Henderson & No & 0.68 & 0.79 & \textbf{0.97} & 0.13 & 0.07\tabularnewline
S. Kums & No & 0.87 & 0.90 & \textbf{0.96} & 0.37 & 0.03\tabularnewline
D. Demme & No & 0.73 & 0.76 & \textbf{0.96} & 0.22 & 0.04\tabularnewline
\midrule
J. Martin & Yes & 0.84 & 0.48 & 0.40 & \textbf{0.98} & 0.21\tabularnewline
Zurutuza & No & 0.90 & 0.75 & 0.39 & \textbf{0.98} & 0.23\tabularnewline
T. Rinc\'{o}n & No & 0.93 & 0.75 & 0.29 & \textbf{0.98} & 0.11\tabularnewline
T. Bakayoko & No & 0.90 & 0.57 & 0.29 & \textbf{0.98} & 0.20\tabularnewline
\midrule
C. Eriksen & Yes & 0.08 & 0.01 & 0.43 & 0.18 & \textbf{0.98}\tabularnewline
J. Pastore & No & 0.12 & 0.02 & 0.56 & 0.31 & \textbf{0.97}\tabularnewline
H. Vanaken & No & 0.31 & 0.03 & 0.58 & 0.35 & \textbf{0.95}\tabularnewline
V. Birsa & No & 0.18 & 0.01 & 0.21 & 0.57 & \textbf{0.94}\tabularnewline
\bottomrule
\end{tabular}
\end{table}

Table~\ref{table:results} shows the predicted probabilties of fulfilling any of the five central midfielder roles for 20 players in our dataset. For each role, the table shows the top-ranked labeled player as well as the top-three-ranked unlabeled players.

The table shows that most players fulfill different types of roles during the course of a season. In general, players who rate high on one role also score high on at least one other role. For example, top-ranked box-to-box midfielders like Jonas Martin (Strasbourg), David Zurutuza (Real Sociedad), Tom\'{a}s Rinc\'{o}n (Torino) and Ti\'{e}mou\'{e} Bakayoko (Chelsea) also score high for the ball-winning midfielder role. The advanced playmaker (AP) role seems to be an exception though. The players who rate high on this role do not rate high on any of the other roles.

Furthermore, the table also shows that our approach is capable of handling players playing in different leagues. With deep-lying playmaker Sven Kums of Anderlecht and advanced playmaker Hans Vanaken of Club Brugge, two players from the smaller Belgian Pro League appear among the top-ranked players for their respective roles. The table lists six players from the Spanish LaLiga, four players from the English Premier League, three players from the French Ligue Un, three players from the Italian Serie A, two players from the German 1. Bundesliga, two players from the Belgian Pro League, and no players from the Dutch Eredivisie.

\section{Related work}
\label{section:related-work}

The task of automatically deriving roles of players during football matches has largely remained unexplored to date. To the best of our knowledge, our proposed approach is the first attempt to identify specific roles of players during matches fully automatic in a data-driven fashion using a supervised learning approach.

In contrast, most of the work in this area focuses on automatically deriving positions and formations from data in an unsupervised setting. Bialkowski et al.~\cite{bialkowski2014win} present an approach for automatically detecting formations from spatio-temporal tracking data collected during football matches. Pappalardo et al.~\cite{pappalardo2018playerank} present a similar approach for deriving player positions from play-by-play event data collected during football matches. Our approach goes beyond their approaches by not only deriving each player's standard position in a formation (e.g., a left winger in a 4-3-3 formation) but also his specific role within that position (e.g., a holding midfielder or a ball-winning midfielder).

Furthermore, researchers have compared football players in terms of their playing styles. Mazurek~\cite{mazurek2018football} investigates which player is most similar to Lionel Messi in terms of 24 advanced match statistics. Pe{\~n}a and Navarro~\cite{pena2015can} analyze passing motifs to find an appropriate replacement for Xavi who exhibits a similar playing style. In addition, several articles discussing playing styles and player roles have recently appeared in the mainstream football media (e.g., \cite{arse2016dm,espn2018dm,thompson2018four}).

\vspace{40pt}

\section{Conclusion}

This paper proposed 21 player roles to characterize the playing styles of football players and presented an approach to automatically derive the most applicable roles for each player from play-by-play event data. Our supervised learning approach proceeds in two steps. First, we compute a large set of statistics and metrics that both characterize the different roles and help distinguish between the roles from match data. Second, we perform a binary classification task for each role leveraging labeled examples obtained from the SciSports Datascouting department. Our approach goes beyond existing techniques by not only deriving each player's standard position (e.g., an attacking midfielder in a 4-2-3-1 formation) but also his specific role within that position (e.g., an advanced playmaker). Our experimental evaluation demonstrates our approach for deriving five roles for central midfielders from data collected during the 2017/2018 season.

In the future, we plan to further improve and extend our approach. We will investigate different learning algorithms to tackle the classification task (e.g.,~\texttt{XGBoost}) as well as different learning settings. More specifically, we aim to obtain a richer set of labels from the SciSports Datascouting department. In addition to each player's primary role, we also wish to collect possible alternative player roles, which would turn our task into a multi-label classification setting.

\clearpage

\bibliographystyle{splncs04}
\bibliography{paper}

\end{document}